\pdfoutput=1
\documentclass[11pt]{article}

\usepackage[margin=1in]{geometry}
\usepackage[T1]{fontenc}
\usepackage{newtxtext,newtxmath}
\usepackage{microtype}
\usepackage{amsmath}
\usepackage{bm}
\usepackage{graphicx}
\usepackage{booktabs}
\usepackage{tabularx}
\usepackage{array}
\usepackage{multirow}
\usepackage{enumitem}
\usepackage[font=small,labelfont=bf,skip=6pt]{caption}
\usepackage{xcolor}
\usepackage{fancyhdr}
\usepackage[hidelinks]{hyperref}
\hypersetup{colorlinks=true,linkcolor=blue!50!black,citecolor=blue!50!black,urlcolor=blue!50!black}

\setlist{itemsep=2pt,topsep=4pt}
\newcommand{\evidence}[1]{\par\smallskip\noindent\textit{\small Evidence class: #1}\par\smallskip}

\fancypagestyle{firstpage}{\fancyhf{}\fancyfoot[C]{\small\thepage}}

\title{\LARGE\bfseries Imitation Learning for Autonomous Driving in CARLA}
\author{%
  \textbf{Jordy Kieto}\\[4pt]
  \small Code, data samples and policy: \url{https://github.com/JordyKieto/carla-imitation-learning}\\
  \small Training curves: \href{https://github.com/JordyKieto/carla-imitation-learning/blob/main/results/training_curves.ipynb}{\texttt{results/training\_curves.ipynb}}
  \quad$\cdot$\quad Video: \url{https://youtu.be/rr_uS4bf0B4}
}
\date{}

\begin{document}

\maketitle
\thispagestyle{firstpage}

\begin{abstract}
\noindent Behavioral cloning trains a policy offline, under the expert's state distribution, but deploys it autoregressively: every action the policy takes shapes the observations it receives next. We study how much closed-loop driving competence a compact multimodal policy can acquire from offline expert demonstrations in the CARLA simulator, and how the design of the demonstration distribution affects what emerges under deployment. The policy fuses 5-frame histories of RGB images, a LiDAR bird's-eye-view histogram, vehicle telemetry and lane waypoints through per-modality encoders and a two-layer temporal Transformer, and regresses continuous throttle, brake and steering at 20\,Hz. Demonstrations went through three stages: a small manually collected set, broader curated recordings, and finally a \emph{systematic route-generation} procedure. That procedure enumerates spawn points, queries the road topology for feasible maneuvers, and keeps autopilot rollouts only after verifying the maneuver actually executed. The released policy (1.36\,M parameters) was trained on 236,882 windows ($\approx$3.3\,h) from 448 such captures. Despite the simplicity of offline behavioral cloning, the resulting policy drives autoregressively, alone on the road and without human intervention, for hours on both training routes and held-out routes. In the author's runs it did so without collisions, and it transfers qualitatively to an unseen CARLA town whose road geometry differs from its demonstration towns. We additionally observe recovery from large trajectory deviations, although we do not claim systematic recovery without controlled evaluation. We report measured offline metrics, reconstruct the provenance of the released checkpoint from logged artifacts, and state explicitly which closed-loop findings are observed rather than measured. All code, the checkpoint (Keras and ONNX), a data sample and a claim-by-claim evidence audit are released.
\end{abstract}

\section{Introduction}

End-to-end driving policies trained by imitation are attractive because they turn a hard control problem into supervised learning: record an expert, then regress its actions from its observations~\cite{dosovitskiy2017carla,bojarski2016end}. The difficulty is that the supervised objective and the deployment setting disagree. Training minimizes error on states the \emph{expert} visited. Deployment runs the \emph{learner} in a loop, where its own small errors move it into states the expert never demonstrated. Those errors can compound~\cite{ross2011dagger}. Low validation loss is therefore necessary but not sufficient evidence of driving ability. The object of interest is the closed-loop behavior.

This report documents a completed imitation-learning experiment built around that distinction. Rather than proposing a new architecture, we use a deliberately compact multimodal temporal policy as an instrument. The question we ask is:

\begin{quote}
\itshape How much closed-loop driving competence can a multimodal behavioral-cloning policy acquire from offline expert demonstrations, and how does the coverage of the demonstration distribution shape the behaviors that emerge under autoregressive deployment?
\end{quote}

The project progressed from a tiny demonstration set to systematic, verified route generation. This amounts to an intervention on the \emph{support} of the demonstration distribution, which is the quantity behavioral cloning is fundamentally limited by.

\paragraph{Contributions.}
\begin{enumerate}
  \item \textbf{A systematic demonstration-generation procedure} for CARLA. It enumerates spawn points, derives the feasible maneuvers at the next junction from the road graph, randomizes start poses, and accepts an autopilot rollout only after classifying the maneuver it actually executed. It produces a maneuver-balanced set of short, reproducible, labeled captures (\S\ref{sec:demos}).
  \item \textbf{A reproducible behavioral-cloning pipeline and released policy.} It covers sensor rig, data contract, preprocessing, window indexing and training, plus the trained checkpoint in Keras and ONNX, with provenance tied to a logged training run (\S\ref{sec:policy}, \S\ref{sec:training}, Appendix~\ref{app:provenance}).
  \item \textbf{An evidence-graded account of closed-loop behavior} (hours of autonomous driving on training and held-out routes, zero-shot transfer to an unseen town, recovery from deviations), with each claim classified as measured, observed, inferred or unverified (\S\ref{sec:closedloop} and the released claim audit).
\end{enumerate}

\section{Related Work}

\paragraph{End-to-end driving from demonstrations.} ALVINN~\cite{pomerleau1989alvinn} and NVIDIA's PilotNet~\cite{bojarski2016end} showed that a network can map camera images directly to steering. PilotNet already recognized the covariate-shift problem and augmented training with synthetically shifted and rotated views to teach recovery. Our policy follows this lineage but is multimodal and temporal, and it deliberately contains \textbf{no synthetic recovery augmentation} (\S\ref{sec:recovery}).

\paragraph{CARLA and conditional imitation.} CARLA~\cite{dosovitskiy2017carla} was introduced with an imitation-learning baseline among its reference agents, so our experiment sits inside one of the simulator's original research purposes. Conditional imitation learning~\cite{codevilla2018cil} resolves the ambiguity of junctions by conditioning the policy on a navigational command. Our policy instead receives a local lane-waypoint path. As we show in \S\ref{sec:obs}, that path does not always disambiguate junctions, which makes this a weaker form of conditioning than~\cite{codevilla2018cil}.

\paragraph{Covariate shift and data aggregation.} Ross and Bagnell~\cite{ross2010efficient} showed that behavioral cloning's cost can grow quadratically with horizon, and DAgger~\cite{ross2011dagger} fixes this by querying the expert on learner-visited states. Prakash et al.~\cite{prakash2020exploring} studied which aggregated states matter for vision-based urban driving in CARLA. DART~\cite{laskey2017dart} injects noise into the \emph{expert's} demonstrations so that recovery states appear in the data. Our route generator's randomized start poses also introduce limited off-nominal states into the demonstrations. This is conceptually related to expert-noise approaches such as DART, although the perturbations were designed for route diversity rather than recovery training (\S\ref{sec:routegen}). We account for it when interpreting recovery.

\paragraph{Privileged teachers.} Learning by Cheating~\cite{chen2019lbc} trains a privileged agent on ground-truth simulator state and distills it into a sensorimotor student. Our expert, the CARLA Traffic Manager autopilot, is also privileged. Unlike~\cite{chen2019lbc}, our student \emph{also} receives privileged route context at test time, so we state explicitly what information crosses that boundary (\S\ref{sec:obs}, \S\ref{sec:limitations}).

\section{Problem Formulation}
\label{sec:problem}

An expert policy $\pi^\ast$ produces a demonstration dataset
\begin{equation}
\mathcal{D} = \{(o_t, a_t)\}_{t=1}^{N}, \qquad a_t = (\tau_t,\ b_t,\ \delta_t) \in [0,1]\times[0,1]\times[-1,1],
\end{equation}
of throttle $\tau$, brake $b$ and steering $\delta$ at 20\,Hz. The policy observes a temporal stack of $K=5$ frames (0.25\,s) across four modalities,
\begin{equation}
o_t = \big(\, I_{t-K+1:t},\ L_{t-K+1:t},\ s_{t-K+1:t},\ W_{t-K+1:t} \,\big),
\end{equation}
where $I$ is RGB, $L$ a LiDAR bird's-eye-view (BEV) histogram, $s$ telemetry and $W$ lane waypoints, all in the ego frame.

\paragraph{Offline objective.} Behavioral cloning fits $\pi_\theta$ by empirical risk minimization under the \emph{expert} state distribution $d_{\pi^\ast}$:
\begin{equation}
\theta^\ast = \arg\min_\theta\ \mathbb{E}_{(o,a)\sim d_{\pi^\ast}} \big[\, \ell(\pi_\theta(o), a)\, \big], \qquad
\ell(\hat a, a) = H_1(\hat\tau-\tau) + H_1(\hat b-b) + (\hat\delta-\delta)^2,
\label{eq:loss}
\end{equation}
with the Huber loss $H_1(x) = \tfrac12 x^2$ for $|x|\le 1$ and $|x|-\tfrac12$ otherwise.

\paragraph{Closed-loop deployment.} At test time, actions feed back into the simulator,
\begin{equation}
o_t \xrightarrow{\ \pi_\theta\ } \hat a_t \xrightarrow{\ \text{CARLA}\ } o_{t+1} \xrightarrow{\ \pi_\theta\ } \hat a_{t+1} \longrightarrow \cdots,
\end{equation}
so the learner is evaluated under its \emph{own} distribution $d_{\pi_\theta} \neq d_{\pi^\ast}$. If $\epsilon$ is the learner's per-step error rate under $d_{\pi^\ast}$, the expected cost over a horizon $T$ is bounded only as~\cite{ross2010efficient}
\begin{equation}
J(\pi_\theta) \;\le\; J(\pi^\ast) + \mathcal{O}(T^2 \epsilon),
\label{eq:bound}
\end{equation}
because a single mistake can place the learner in states with no demonstrations. At 20\,Hz a one-minute drive is $T=1200$ decisions. Held-out action error measures only $\epsilon$ under expert states; it does not by itself establish driving competence, which must ultimately be assessed closed-loop (e.g.\ by route completion and infractions, as in CARLA's own benchmarks~\cite{dosovitskiy2017carla}).

\section{Multimodal Temporal Policy}
\label{sec:policy}

\begin{figure}[!htbp]
  \centering
  \includegraphics[width=0.78\linewidth]{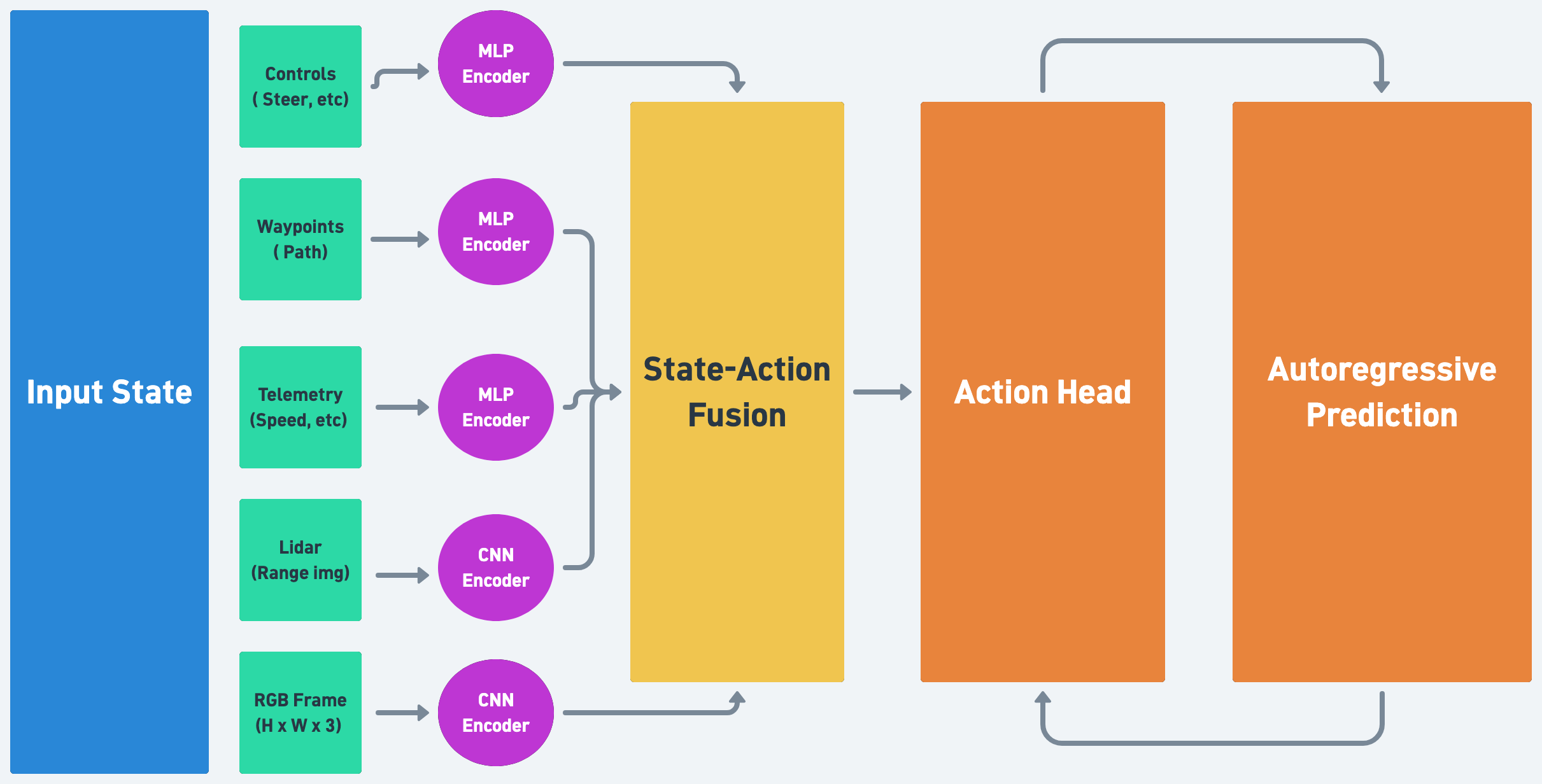}
  \caption{Conceptual overview of the policy. The released checkpoint's inputs are RGB, LiDAR BEV, telemetry and waypoints; past controls are not an input (\S\ref{sec:arch}).}
  \label{fig:arch}
\end{figure}

\subsection{Observations}
\label{sec:obs}

\begin{table}[!htbp]
  \centering
  \small
  \caption{Observation modalities.}
  \label{tab:obs}
  \begin{tabularx}{\linewidth}{@{}l X l@{}}
    \toprule
    Modality & Construction & Shape \\
    \midrule
    RGB $I_t$ & Forward camera, FOV $90^\circ$, mounted at $(0.4, 0, 1.6)$\,m; captured at $224^2$, bilinearly resized to $96^2$, scaled to $[0,1]$ & $96\times96\times3$ \\
    LiDAR $L_t$ & 32-channel ray-cast LiDAR, 50\,m range, 50k pts/s; BEV histogram (Eq.~\ref{eq:bev}) & $64\times64\times1$ \\
    Telemetry $s_t$ & Soft-saturated kinematics (Eq.~\ref{eq:telemetry}) & $5$ \\
    Waypoints $W_t$ & 25 lane-center waypoints spaced 2\,m (Eq.~\ref{eq:waypoints}) & $25\times3$ \\
    \bottomrule
  \end{tabularx}
\end{table}

\paragraph{LiDAR BEV.} For points $(x_i, y_i)$ in the sensor frame with $0 < x_i < 30$\,m and $|y_i| < 10$\,m,
\begin{equation}
H_{jk} = \sum_i \mathbb{1}\!\left[(x_i, y_i)\in \mathrm{bin}_{jk}\right], \qquad L_t = \frac{H}{\max_{jk} H_{jk} + 10^{-6}} .
\label{eq:bev}
\end{equation}

\paragraph{Telemetry.} With world-frame velocity $v$, acceleration $\alpha$ and yaw $\psi$, forward and lateral speeds are $v_f = v_x\cos\psi + v_y\sin\psi$ and $v_\ell = -v_x\sin\psi + v_y\cos\psi$, and
\begin{equation}
s_t = \Big(\tanh\tfrac{\lVert v\rVert}{20},\ \tanh\tfrac{3.6\lVert v\rVert}{72},\ \tanh\tfrac{\lVert \alpha\rVert}{5},\ \tanh\tfrac{v_f}{20},\ \tanh\tfrac{v_\ell}{20}\Big).
\label{eq:telemetry}
\end{equation}
The first two components are identical by construction ($3.6/72 = 1/20$), so the policy effectively receives four distinct telemetry signals.

\paragraph{Waypoints (route context).} Starting from the lane waypoint nearest the ego vehicle, successive waypoints $w_{k+1} = \mathrm{next}_{2\,\mathrm{m}}(w_k)$ are expressed in the ego frame, $p_k = T_{\text{ego}}^{-1} w_k = (x_k, y_k)$, with $d_k = \lVert p_k\rVert$, and encoded as
\begin{equation}
W_t[k] = \Big(\tfrac{x_k}{d_k},\ \tfrac{y_k}{d_k},\ \tfrac{\min(d_k, 50)}{50}\Big), \qquad k = 0,\dots,24 .
\label{eq:waypoints}
\end{equation}
Two properties matter for interpretation. (i) Waypoints come from the simulator's HD-map lane graph and the \textbf{ground-truth ego pose}, which makes them privileged information. (ii) At a junction $\mathrm{next}(\cdot)$ has several successors and the generator takes the first. The waypoint path therefore follows the lane graph rather than the expert's chosen maneuver, and can disagree with it before a turn.

\begin{figure}[!htbp]
  \centering
  \includegraphics[width=0.72\linewidth]{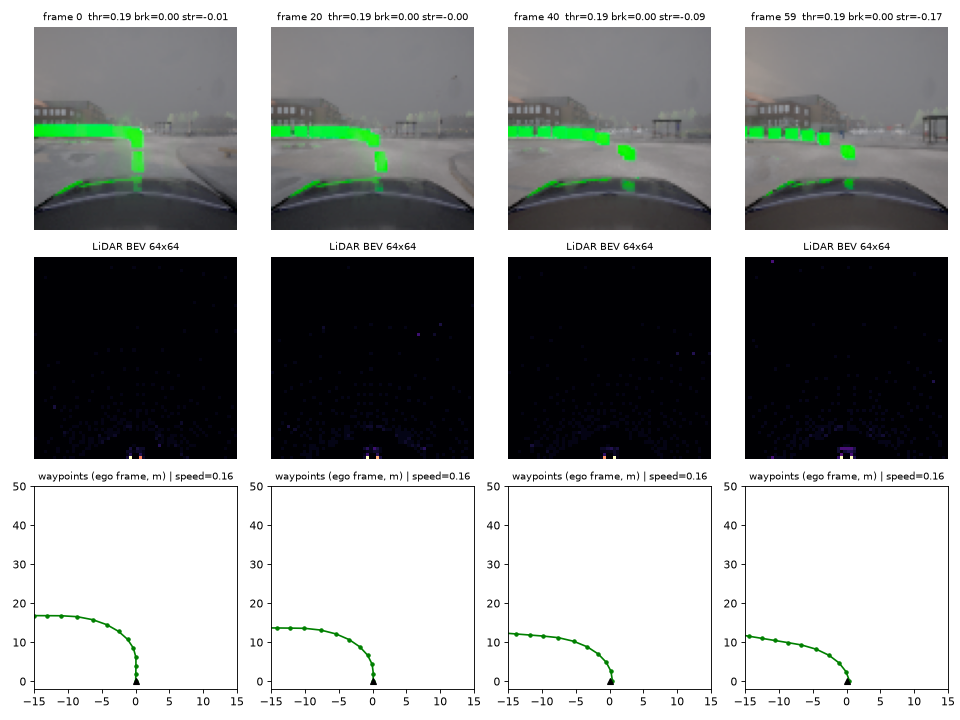}
  \caption{Four frames (3\,s) of processed training data entering a left turn: RGB (top), LiDAR BEV (middle), and the waypoint path in metres in the ego frame (bottom). The green points in the RGB frames are debug waypoints drawn in the simulator during collection; they are \emph{visible to the camera} (\S\ref{sec:limitations}).}
  \label{fig:obs}
\end{figure}

\subsection{Architecture}
\label{sec:arch}

Per timestep, each modality has its own encoder with weights shared across time:
\begin{equation}
z_t = \big[\, f_{\text{rgb}}(I_t);\ f_{\text{lidar}}(L_t);\ g_{w}(W_t);\ g_{s}(s_t) \,\big] \in \mathbb{R}^{128+128+32+32},
\end{equation}
where $f_{\text{rgb}}$ and $f_{\text{lidar}}$ are three strided $3{\times}3$ convolutions (32, 64, 128 channels, ReLU) with global average pooling, and $g_w$, $g_s$ are single dense layers (32 units, ReLU). The fused features are projected and position-encoded,
\begin{equation}
h^{(0)}_t = \mathrm{ReLU}(W_p z_t) + e_t \in \mathbb{R}^{256},
\end{equation}
then processed by two post-norm Transformer blocks~\cite{vaswani2017attention} (4 heads, key dimension 64, feed-forward 512):
\begin{equation}
\tilde h = \mathrm{LN}\big(h + \mathrm{MHA}(h,h,h)\big), \qquad h' = \mathrm{LN}\big(\tilde h + W_2\,\mathrm{ReLU}(W_1 \tilde h)\big).
\end{equation}
The action is decoded from the most recent timestep, $\hat a_t = W_o\,\mathrm{ReLU}(W_h\, h^{(2)}_t)$. The model has \textbf{1,358,211 parameters}. At inference, throttle and brake are clipped to $[0,1]$ and steering to $[-1,1]$.

\paragraph{Positional embedding.} Inspecting the serialized checkpoint shows that $e_t$ is not a trainable parameter. The embedding layer was applied to a constant index tensor, so the framework evaluated it once at construction and stored its randomly initialized output as a constant. Temporal order is therefore still encoded, but by a fixed random code rather than a learned one. We report the model as it was actually trained.

We do not claim architectural novelty. The design is a compact, conventional late-fusion temporal policy chosen so that the scientific variables are the data and the closed-loop behavior.

\section{Demonstrations}
\label{sec:demos}

\subsection{Expert and label construction}

The expert is either a human driver (keyboard) or the CARLA Traffic Manager autopilot, which acts on privileged simulator state. The simulator runs in synchronous mode at $\Delta t = 0.05$\,s. Camera and LiDAR messages are paired only if their frame indices differ by at most one and their timestamps by at most $1.5\,\Delta t$. Otherwise the frame is dropped rather than recorded misaligned.

Keyboard control is binary, so recorded labels are conditioned before storage. Throttle and steering pass through an exponential low-pass filter and a rate limiter:
\begin{align}
\bar u_t &= \alpha\, u_t + (1-\alpha)\,\bar u_{t-1}, \qquad \alpha = \frac{\Delta t}{\tau_u + \Delta t}, \\
u^{\text{label}}_t &= u^{\text{label}}_{t-1} + \mathrm{clip}\big(\bar u_t - u^{\text{label}}_{t-1},\ -r^{\downarrow}_u\Delta t,\ r^{\uparrow}_u \Delta t\big),
\end{align}
with $\tau_{\text{throttle}} = 0.5$\,s, $r^{\uparrow} = 1.0$/s, $r^{\downarrow} = 1.5$/s and a 0.03 deadband, and $\tau_{\text{steer}} = 0.2$\,s with $r = 2.5$/s. Brake is recorded raw. The learning target is therefore a smooth, physically plausible control signal rather than a bang-bang keypress.

\subsection{From small demonstrations to systematic route generation}
\label{sec:routegen}

The demonstration set evolved in three stages.
\begin{enumerate}
  \item \textbf{Minimal manual demonstrations.} Early experiments used a very small set of manual demonstrations. The author reports that roughly seven minutes of driving already produced a policy that followed the road in Town01, with about one in ten rollouts ending in a collision. \emph{These figures are author-reported; no run or dataset from this stage is part of the released artifact.} The qualitative lesson that drove the next stages was that recognizable closed-loop behavior can emerge from very little data, but the failures cluster where the data is thin.
  \item \textbf{Broader curated demonstrations} extended coverage to hours of driving.
  \item \textbf{Systematic route generation} (\texttt{collect.py}, first committed 2026-05-05) replaced ad hoc collection with a controlled procedure. This procedure produced the released policy's training data.
\end{enumerate}

\paragraph{Route generation.} Autopilot demonstrations have a subtle flaw: at an intersection the autopilot's choice is stochastic, so turns are under-represented and uncontrolled. The generator makes coverage a design variable:
\begin{itemize}
  \item \textbf{Enumerate} every spawn point $p$ on the map.
  \item \textbf{Query feasible maneuvers.} Walk the lane graph forward in 2\,m steps (up to 40) to the first branching point. For each successor branch, compare the heading 16\,m further along with the heading at the branch, $\Delta\psi = \mathrm{wrap}(\psi_{\text{branch}} - \psi_{\text{ref}})$, and label the maneuver $m = \texttt{left}$ if $\Delta\psi > 25^\circ$, $\texttt{right}$ if $\Delta\psi < -25^\circ$, else $\texttt{straight}$.
  \item \textbf{Randomize the start.} Sample $\Delta\psi_0\sim\mathcal U(-3^\circ, 3^\circ)$, lateral offset $\sim\mathcal U(-0.25, 0.25)$\,m and longitudinal offset $\sim\mathcal U(-0.5, 0.5)$\,m.
  \item \textbf{Roll out} the autopilot for 30\,s (600 frames) at a fixed 18\,mph target speed, with lane changes disabled.
  \item \textbf{Verify, then accept.} Classify the executed maneuver from the ego yaw trajectory,
  \[ \hat m = \mathrm{cls}\Big(\max_t \big|\mathrm{wrap}(\psi_t - \psi_0)\big|\Big), \]
  with the same $25^\circ$ threshold applied to the signed extremum. Keep the capture only if $\hat m$ is feasible at this spawn point and its quota (captures per maneuver) is not yet filled. Retry up to 20 times per spawn point.
\end{itemize}

The verify-then-accept step turns a stochastic expert into a \emph{controlled} data source. Every stored capture carries a verified maneuver label, the spawn index, the start-pose perturbation and sensor-alignment statistics. The randomized start also has a second-order effect: each capture begins slightly off the lane center and heading. This introduces limited off-nominal states, conceptually related to expert-noise approaches such as DART~\cite{laskey2017dart}, although the perturbations were designed for route diversity rather than recovery training. We return to this in \S\ref{sec:recovery}.

\subsection{The training set}
\label{sec:trainset}

We reconstructed the released policy's training set from logged artifacts (Appendix~\ref{app:provenance}); Table~\ref{tab:data} summarizes it.

\begin{table}[!htbp]
  \centering
  \small
  \caption{Training set of the released policy.}
  \label{tab:data}
  \begin{tabular}{@{}ll@{}}
    \toprule
    Quantity & Value \\
    \midrule
    Maps & \textbf{Town01, Town02} \\
    Recordings & \textbf{448} (394 on 2026-05-09, 54 on 2026-05-10) \\
    Median recording length & 595 windows $\approx$ 600 frames $=$ 30\,s (route generator capture length) \\
    Five-frame windows & \textbf{236,882} $\approx$ \textbf{3.3\,h} of 20\,Hz driving \\
    Direction mix (by window) & straight 89.47\%, left 5.85\%, right 4.67\% \\
    Windows with brake $> 0$ & 2.0\% \\
    Mean target throttle / mean $|\delta|$ & 0.235 / 0.030 \\
    \bottomrule
  \end{tabular}
\end{table}

\paragraph{Windowing.} Each frame gets a direction label, \texttt{left} if $\delta \le -0.08$, \texttt{right} if $\delta \ge 0.08$, \texttt{straight} otherwise. A window $[t-4, t]$ is kept only if its majority direction covers at least 60\% of its frames. This filter removes ambiguous transition windows so that the per-direction diagnostics in \S\ref{sec:training} are well defined. It also means some turn-entry transitions are excluded from training.

Even with maneuver-balanced \emph{captures}, the balance by \emph{window} is dominated by straight driving (89\%), because a 30\,s capture spends most of its frames approaching and leaving the junction. This is a concrete example of why demonstration coverage has to be measured at the unit the loss actually sees.

\section{Training and Offline Evaluation}
\label{sec:training}

\subsection{Protocol}

The model was trained with Adam (learning rate $3\times10^{-3}$, batch size 409) on the loss of Eq.~\ref{eq:loss}. Windows were split \emph{per recording and per direction}: the first 70\% of each recording's windows of a given direction went to training and the remaining 30\% to validation, giving 165,777 training and 71,108 validation windows with matching direction ratios. Every 50 steps the model was evaluated on 50 validation batches. The run lasted 53 epochs (21,750 steps, 8.88\,M windows, 1.64\,h), and checkpoints were written every 10 epochs. \textbf{The released policy is the epoch-10 checkpoint.}

This split is a \emph{temporal} hold-out within recordings. It measures interpolation to unseen moments of seen routes, not generalization to unseen routes or maps. We therefore use it only to diagnose fitting, never as evidence of driving ability.

\subsection{Metrics}

Beyond per-control losses and MAE, we logged diagnostics aimed at driving-relevant failure modes:
\begin{itemize}
  \item \textbf{Turn detection.} With $\mathcal P = \{t : |\hat\delta_t| > 0.1\}$ and $\mathcal G = \{t : |\delta_t| > 0.1\}$, we report precision $|\mathcal P\cap\mathcal G|/|\mathcal P|$, recall $|\mathcal P\cap\mathcal G|/|\mathcal G|$ and their harmonic mean $F_1$.
  \item \textbf{Turn onset delay} $\min\mathcal P - \min\mathcal G$, in frames.
  \item \textbf{Lagged error} $\tfrac{1}{n-k}\sum_t|\hat a_t - a_{t+k}|$ for $k\in\{1,2,3\}$, the error against targets $k$ frames in the future. Error that shrinks with $k$ would indicate anticipation of upcoming controls; error that grows with $k$ indicates predictions tracking the current control.
  \item \textbf{Smoothness} $\tfrac{1}{n-1}\sum_t |\hat a_{t+1}-\hat a_t|$.
  \item \textbf{Per-direction MSE}, to expose the class imbalance of \S\ref{sec:trainset}.
\end{itemize}

\subsection{Results}

\begin{figure}[!htbp]
  \centering
  \includegraphics[width=0.95\linewidth]{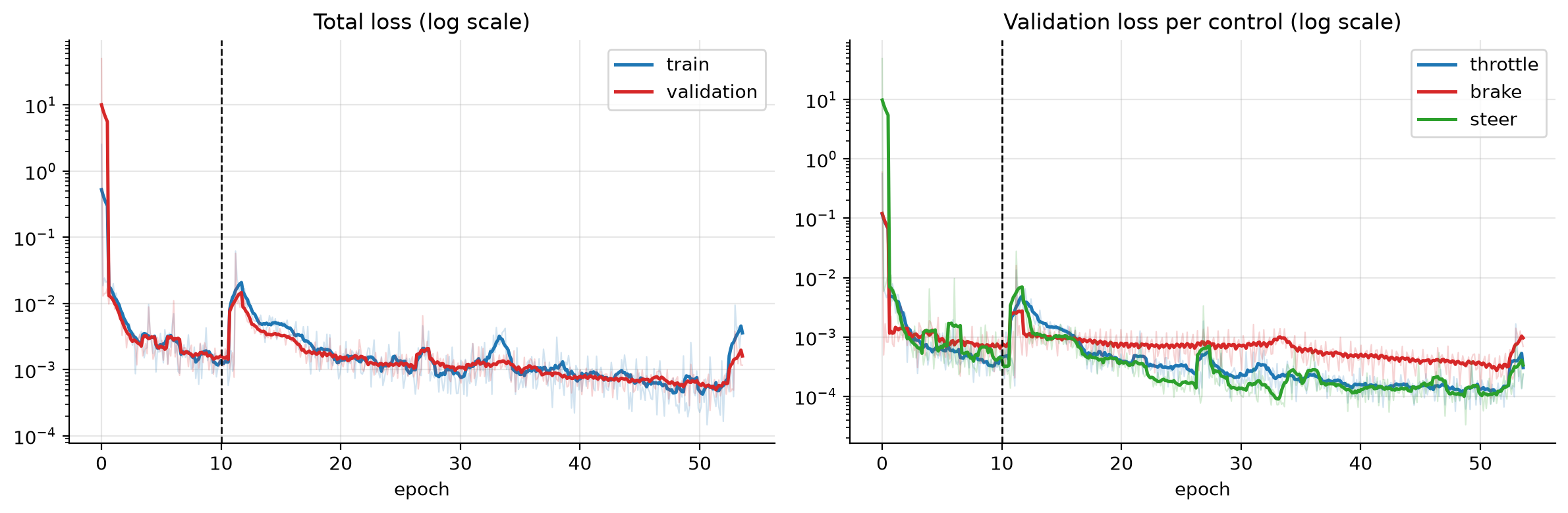}\\[4pt]
  \includegraphics[width=\linewidth]{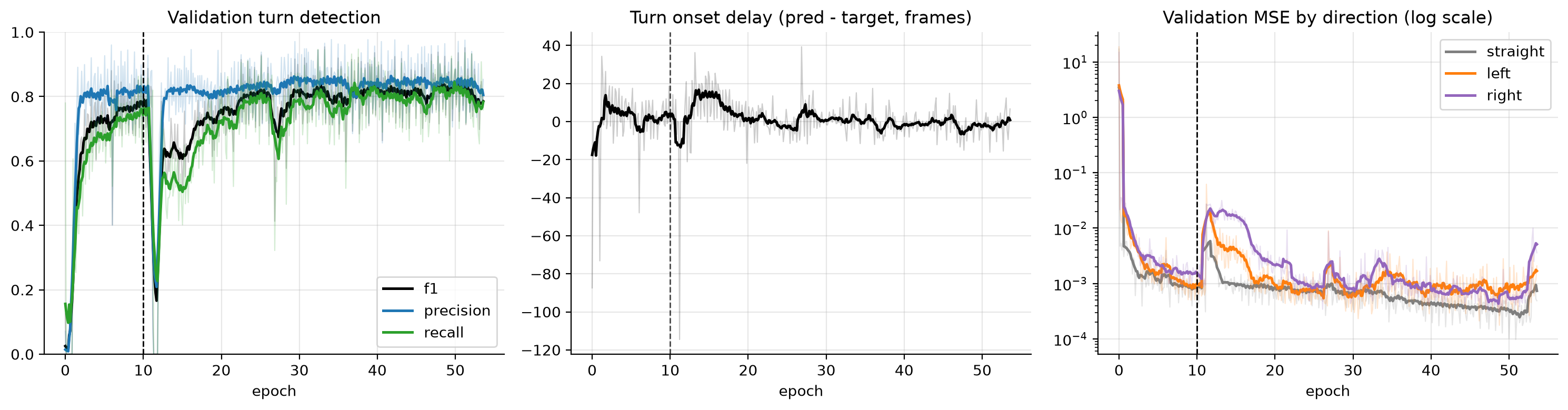}
  \caption{Training curves of the full run, regenerated from the exported metric history. Top: total train/validation loss and per-control validation loss. Bottom: validation turn detection, turn onset delay and per-direction validation MSE. Faint lines are raw evaluations, solid lines a 9-point moving average; the dashed line marks the released epoch-10 checkpoint. Source: \texttt{results/training\_curves.ipynb}.}
  \label{fig:curves}
\end{figure}

\begin{table}[!htbp]
  \centering
  \small
  \setlength{\tabcolsep}{4.5pt}
  \caption{Offline validation metrics. Epoch 1: last evaluation of the epoch. Other rows: mean over all validation evaluations within the stated epochs.}
  \label{tab:offline}
  \begin{tabular}{@{}lcccccccc@{}}
    \toprule
    & & \multicolumn{3}{c}{MAE} & & \multicolumn{3}{c}{Val MSE by direction} \\
    \cmidrule(lr){3-5}\cmidrule(l){7-9}
    & Val loss & throttle & brake & steer & Turn $F_1$ & straight & left & right \\
    \midrule
    Epoch 1 (end) & 1.26e-2 & 0.072 & 0.009 & 0.056 & 0.39 & 5.8e-3 & 9.3e-3 & 1.3e-2 \\
    \textbf{Epoch 10 (released)} & \textbf{1.53e-3} & \textbf{0.017} & \textbf{0.017} & \textbf{0.013} & \textbf{0.78} & \textbf{0.84e-3} & \textbf{0.98e-3} & \textbf{1.57e-3} \\
    Epochs 50--53 & 0.70e-3 & 0.008 & 0.004 & 0.006 & 0.82 & 0.36e-3 & 0.93e-3 & 0.83e-3 \\
    \bottomrule
  \end{tabular}
\end{table}

Figure~\ref{fig:curves} and Table~\ref{tab:offline} support four observations.
\begin{enumerate}
  \item \textbf{Most of the offline fit happens early.} From epoch 10 to epoch $\geq 50$ the validation loss roughly halves again. Turn $F_1$, the metric closest to a driving decision, only moves from 0.78 to 0.82.
  \item \textbf{Turns remain the hard case.} Throughout training, right-turn validation error stays about 2$\times$ straight-driving error, and by epochs 50--53 left-turn error is about 2.6$\times$ straight. This matches the 5--6\% share of turn windows, and it is exactly where closed-loop failures would be expected.
  \item \textbf{Per-batch $R^2$ is uninformative here.} Logged $R^2$ values are extremely negative (e.g.\ $-3.5\times10^{3}$ for steering at epoch 10) even though MAE is around 0.01. Validation batches are contiguous slices of recordings, so the target variance \emph{within} a batch is often near zero and $R^2 = 1 - \mathrm{SS}_{\text{res}}/\mathrm{SS}_{\text{tot}}$ blows up. We report this so the metric is not misread, and recommend dataset-level $R^2$ in future runs.
  \item \textbf{Training briefly destabilized right after the released checkpoint.} Shortly after epoch 10, train and validation loss spike by more than an order of magnitude (validation loss peaks at $5.7\times10^{-2}$, 37$\times$ the epoch-10 mean) and turn $F_1$ collapses to 0, before smoothed validation loss and turn $F_1$ return to their epoch-10 levels around epochs 20 and 22 (Figure~\ref{fig:curves}). The released policy therefore precedes this episode. Its cause was not investigated.
\end{enumerate}

Offline, a steering MAE of $\approx$0.013 means the policy predicts the expert's steering closely \emph{on expert states}. By \S\ref{sec:problem}, this does not by itself say the policy can drive.

\section{Closed-Loop Behavior}
\label{sec:closedloop}

\subsection{Deployment protocol}

The policy runs at 20\,Hz in synchronous mode. At each tick the latest aligned sensor frame is encoded, pushed into a 5-frame buffer (padded by repetition after a reset), and passed to the policy. The clipped action is then applied and the simulator is advanced. No expert, action filtering or safety controller intervenes. The policy's actions alone determine its future observations. Evaluation maps, spawn points and traffic settings are selected in the driving tool (\texttt{drive.py}).

\subsection{Sustained autoregressive driving}

Here \emph{autoregressive} means the policy alone controls the vehicle: it is the only agent on the road, no expert or human intervenes, and every observation it receives is a consequence of its own previous actions. Under these conditions the released policy drove for \textbf{hours}, on both \textbf{training routes and held-out routes}, following lanes and negotiating turns, with \textbf{no collisions and no human intervention} in the author's runs (\href{https://youtu.be/rr_uS4bf0B4}{video}). At 20\,Hz an hour is $T = 72{,}000$ consecutive self-conditioned decisions. Within this traffic-free setting, lane following and turning were effectively solved at the level of observation.

These runs were not logged as a benchmark (no fixed route set, route-completion or infraction counts), so this is a strong observation rather than a measurement. It also says nothing about traffic, pedestrians or traffic rules, which were absent.

\evidence{qualitative observation (author-run, hours of driving); not yet quantified.}

\begin{table}[!htbp]
  \centering
  \small
  \caption{Proposed protocol for quantitative closed-loop evaluation (not yet run). Each condition would use a fixed set of spawn points and routes, several seeds, and no human intervention.}
  \label{tab:protocol}
  \begin{tabularx}{\linewidth}{@{}l l X@{}}
    \toprule
    Condition & Maps & Metrics \\
    \midrule
    In-distribution & Town01, Town02 & route completion, collisions per km, off-road events per km, interventions \\
    Unseen towns & any town other than Town01/Town02 & same as above \\
    Recovery from offsets & all of the above & recovery rate and time-to-recover (\S\ref{sec:recovery}) \\
    \bottomrule
  \end{tabularx}
\end{table}

\subsection{Driving on unfamiliar road geometry}

\begin{figure}[!htbp]
  \centering
  \includegraphics[width=0.7\linewidth]{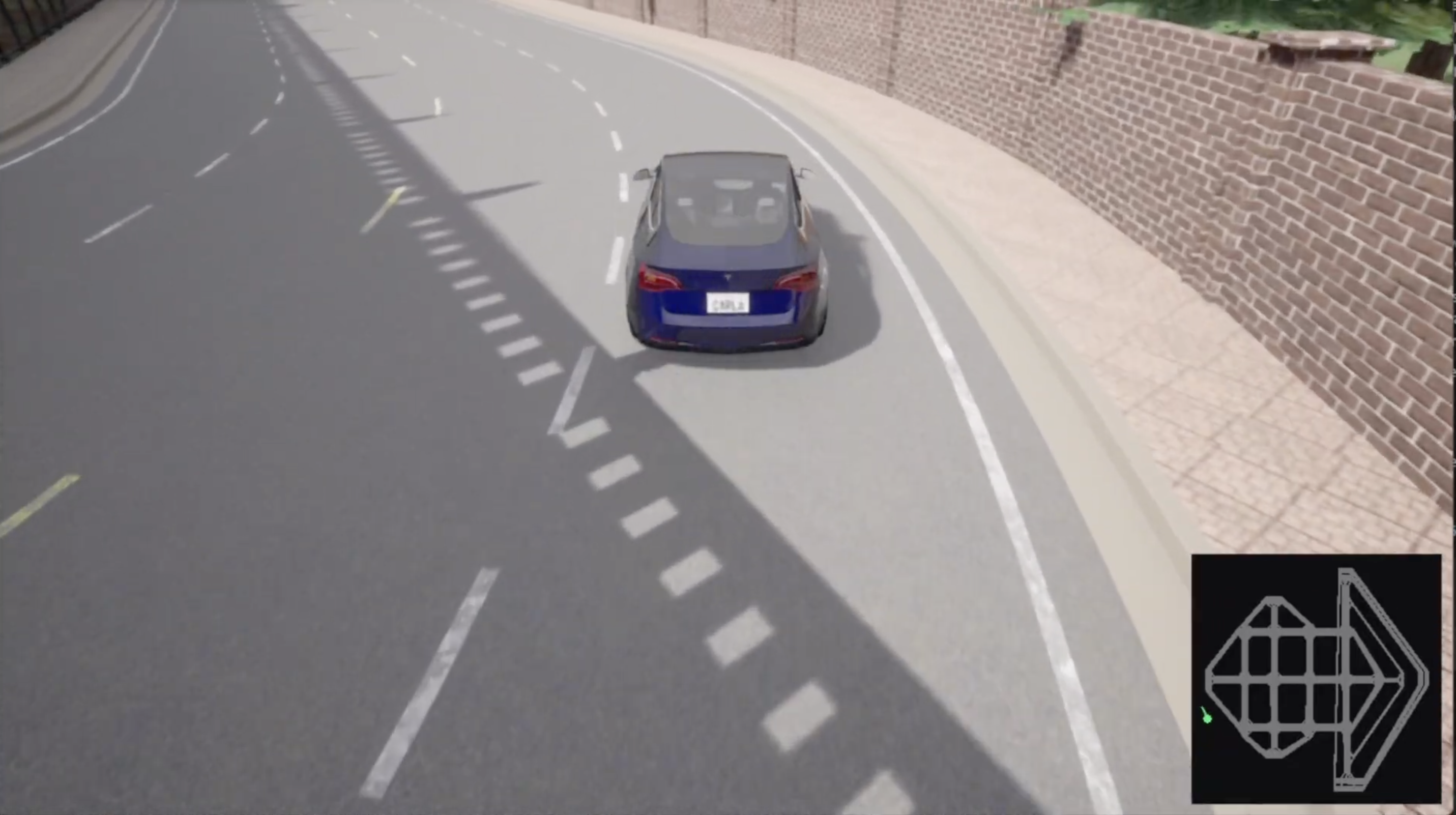}
  \caption{The policy driving a curved multi-lane road in an unseen CARLA town. All demonstrations were collected on Town01 and Town02. The minimap (bottom right) shows the map layout.}
  \label{fig:unseen}
\end{figure}

All demonstrations were collected on Town01 and Town02. Both are small towns whose roads are two-lane streets with T-junctions. Figure~\ref{fig:unseen} shows the policy in a town outside that set, driving a multi-lane curved road with different markings, curvature and surroundings. No demonstrations were recorded in this town, so this is an instance of \textbf{zero-shot transfer to an unseen CARLA town}.

A plausible interpretation is that the policy learned a \emph{local} conditional driving function,
\begin{equation}
\pi\big(a_t \mid \text{visual road geometry},\ \text{local occupancy},\ \text{ego kinematics},\ \text{local route}\big),
\end{equation}
rather than memorizing map-specific appearance or positions: none of its inputs encodes global position or map identity. We state this as an interpretation, not a demonstrated mechanism. It is also consistent with the policy relying heavily on the privileged waypoint path, which is available on any map (\S\ref{sec:limitations}). Separating these explanations requires input ablations (\S\ref{sec:future}).

\evidence{qualitative observation in an unseen town; not yet a systematic measurement (Table~\ref{tab:protocol}).}

\subsection{Recovery from large deviations}
\label{sec:recovery}

In behavioral cloning the textbook failure is a drift into a state the expert never visited, followed by further errors. During closed-loop runs the author observed the policy leaving the lane or road, steering back toward drivable surface, reacquiring the lane and resuming route following.

\paragraph{Recovery evaluation protocol (proposed).} Recovery can be measured directly by initializing the policy at a controlled offset from the lane center. Given a lane waypoint with pose $(x_0, y_0, \psi_0)$, spawn the vehicle at
\begin{equation}
(x, y, \psi) = \big(x_0 - \Delta y\,\sin\psi_0,\ \ y_0 + \Delta y\,\cos\psi_0,\ \ \psi_0 + \Delta\psi\big),
\end{equation}
with lateral offsets $\Delta y \in \{\pm1, \pm2, \pm3\}$\,m and heading offsets $\Delta\psi \in \{0^\circ, \pm15^\circ, \pm30^\circ\}$, i.e.\ up to $12\times$ the lateral and $10\times$ the heading perturbation present in the demonstrations (\S\ref{sec:routegen}). A rollout counts as a \textbf{recovery} if, within $T_r = 5$\,s and without collision, the vehicle reaches $|e_y| < 0.5$\,m and $|e_\psi| < 5^\circ$ relative to the lane and holds this for 1\,s. Reporting the recovery rate and time-to-recover as a function of $(\Delta y, \Delta\psi)$ turns the qualitative observation into a curve showing where recovery breaks down.

\paragraph{What the data rules out, and what it does not.} The development pipeline later gained a dedicated \emph{recovery-route} generator. We verified that its data (4,350 windows, recorded 2026-05-17) was collected \emph{after} the released checkpoint (2026-05-10) and is \textbf{not} in the training set. No online DAgger recordings are in the training set either. The observed recoveries are therefore not imitated from recovery or learner-state demonstrations. The one remaining source of off-nominal states is the route generator's randomized start poses ($\pm$0.25\,m, $\pm3^\circ$), which begin every capture slightly perturbed, although far smaller than the deviations recovered from. A natural hypothesis is that the autopilot's corrections from these small perturbations, combined with the LiDAR and waypoint geometry, generalize to larger deviations. That is an inference to be tested, not a finding.

\evidence{qualitative observation; a recovery mechanism is not claimed. The protocol above would quantify it.}

\section{Limitations and Threats to Validity}
\label{sec:limitations}

\begin{itemize}
  \item \textbf{Evidence strength.} Closed-loop results (hours of intervention-free driving, unseen-town transfer, recovery) are author observations, not logged benchmark metrics. There are no multi-seed retrainings, no systematic rollout statistics (Table~\ref{tab:protocol} is a protocol, not results) and no architecture or input ablations.
  \item \textbf{Privileged inputs at test time.} Waypoints come from the HD-map lane graph and the ground-truth ego pose. In addition, debug waypoints drawn during collection (and during autopilot/policy driving) are \textbf{rendered into the RGB images}, so route information also reaches the vision channel. We make no claim that the student is free of privileged information.
  \item \textbf{Junction ambiguity.} The waypoint path takes the first lane-graph successor, so before a turn it can contradict the expert's maneuver. Turn behavior is therefore not cleanly conditioned on intent.
  \item \textbf{Evaluation split.} Validation is a temporal hold-out within the training recordings.
  \item \textbf{Data composition.} 89\% straight driving by window. There is no traffic, no pedestrians and no weather variation, and the autopilot ignored lights and signs during route generation. Demonstrations cover only two maps (Town01, Town02). Map tags were recorded for only $\approx$36\% of windows in the manifest, so per-map data proportions are unknown.
  \item \textbf{Implementation details affecting interpretation.} The positional embedding is a fixed random constant (\S\ref{sec:arch}); two telemetry components are duplicates; LiDAR BEV max-normalization is dominated by near-ego returns.
  \item \textbf{Historical reconstruction.} Early-stage quantities ($\approx$7\,min, $\approx$1/10 collisions) are author-reported and not backed by released artifacts.
  \item \textbf{Simulation only.} No claims are made about real-world driving.
\end{itemize}

\section{Future Work}
\label{sec:future}

\begin{enumerate}
  \item \textbf{Systematic closed-loop evaluation} (Table~\ref{tab:protocol}): fixed spawn/route sets on Town01/Town02 and unseen towns, several seeds, route completion, collisions and off-road events per km.
  \item \textbf{Input ablations} to separate representation from privileged context: drop waypoints, remove debug rendering from RGB, and swap in noisy or lagged route context.
  \item \textbf{Demonstration-coverage scaling:} train on 7\,min, 30\,min, 1\,h and 3.3\,h subsets of the verified route captures and measure closed-loop metrics against data size, turning the project's historical progression into a controlled experiment.
  \item \textbf{Recovery attribution:} run the offset-initialization protocol (\S\ref{sec:recovery}) on policies trained with no start jitter, the current jitter and explicit recovery routes, and compare their recovery curves.
  \item \textbf{Intent conditioning:} feed the planned route (or a command, as in~\cite{codevilla2018cil}) instead of first-successor lane waypoints.
\end{enumerate}

\section{Conclusion}

A compact multimodal policy trained purely offline on $\approx$3.3 hours of systematically generated, maneuver-verified autopilot demonstrations drives autoregressively in CARLA, alone on the road, for hours on training and held-out routes without collisions or human intervention. It also drives in an unseen CARLA town and sometimes returns to the road after large deviations, even though no recovery demonstrations were in its training data. Offline metrics show that the policy fits expert actions closely, but under autoregressive deployment that fit is only a precondition. The contribution of this report is less the network than the method: controlling and verifying the demonstration distribution, reconstructing the provenance of the released policy from logged artifacts, and grading every claim by its evidence. These observations motivate a controlled study of how demonstration coverage and privileged route context shape closed-loop competence in behavioral cloning.

The systematic demonstration pipeline emerged after earlier attempts with reinforcement learning and less structured imitation pipelines failed to produce reliable closed-loop driving. Those experiments are outside the evidentiary scope of this report.

\bibliographystyle{unsrt}

\appendix
\section{Artifact Provenance}
\label{app:provenance}

{\small\noindent\begin{tabularx}{\linewidth}{@{}l X@{}}
    \toprule
    Evidence & Value \\
    \midrule
    Training run & \texttt{barzksni} (``MLP -- Generalized Cross Map''), created 2026-05-10 22:00:32 UTC; full metric history in \texttt{results/barzksni/}, plotted in \texttt{results/training\_curves.ipynb} \\
    Checkpoint & \texttt{artifacts/policy.keras}, Keras 3.13.2, \texttt{date\_saved} 2026-05-10 22:19:15 UTC \\
    Link & Epoch 10 completed at 1,115\,s $=$ 22:19:07 UTC; checkpoint saved 8\,s later; checkpoints every 10 epochs \\
    Training maps & Town01, Town02 \\
    Training data & 448 recordings dated 2026-05-09/10; 236,882 windows (training-log sampling summary $=$ manifest subset, identical direction ratios) \\
    Excluded data & \texttt{recovery}-tagged recordings (2026-05-17) postdate the checkpoint; no DAgger-mode recordings \\
    Code timeline & MLP policy (2026-01-27) $\rightarrow$ sensor-sync and data-contract fixes (2026-03-25) $\rightarrow$ first end-to-end driving (2026-04-07) $\rightarrow$ route generator (2026-05-05) $\rightarrow$ training (2026-05-10) $\rightarrow$ ``all maps'' model milestone (2026-05-12) \\
    ONNX & \texttt{artifacts/policy.onnx}, max absolute deviation from Keras $< 4\times10^{-7}$ \\
    \bottomrule
  \end{tabularx}\par}

\end{document}